\documentclass{esannV2}
\usepackage{graphicx}
\usepackage[latin1]{inputenc}
\usepackage{hyperref}
\usepackage{amssymb,amsmath,array}
\usepackage{caption}
\usepackage{subcaption}

\begin{document}
\title{Exact ICL maximization in a non-stationary time extension of the latent block model for dynamic networks}

\author{Marco Corneli, Pierre Latouche and Fabrice Rossi 
%
%
\vspace{.3cm}\\
%
 Universit\'e Paris 1 Panth\'eon-Sorbonne - Laboratoire SAMM \\
90 rue de Tolbiac, F-75634 Paris Cedex 13 - France
}

\newtheorem{remark}{Remark}[section]

\maketitle

\begin{abstract}
The latent block model (LBM) is a flexible probabilistic tool to describe 
interactions between node sets in bipartite networks, but it does not
account for interactions of time varying intensity between nodes in unknown 
classes. In this paper we propose a non stationary temporal extension of the 
LBM that clusters simultaneously the two node sets of 
a bipartite network and constructs classes of time intervals on which interactions are stationary. 
The number of clusters as well as the membership to classes are obtained
by maximizing the exact complete-data integrated likelihood relying on a greedy
search approach. Experiments on simulated and real data 
are carried out in order to assess the proposed methodology.
\end{abstract}

\section{Introduction}
Since the interactions between nodes of a network generally have a time varying intensity, the network has a non trivial time structure that we aim at inferring. The approach we follow to introduce a temporal dimension, consists in partitioning the entire time horizon, during which we observe interactions, in disjoint time intervals, having an arbitrary fixed length. Then, we simultaneously cluster the nodes of the bipartite network and these time intervals, assuming interactions are generated by a latent block model. A similar view is adopted by Randriamanamihaga, C\^{o}me and Govaert \cite{Govaert}, nonetheless with a substantial difference: they consider time intervals whose membership is not hidden but known in advance and hence exogenous, whereas in this paper we infer the membership of each interval by maximizing a likelihood criterion. A similar task is accomplished by Guigour\`es, Boull\'e and Rossi \cite{Rossi} in a different point of view: they do not consider fixed-length time intervals, but associate a time stamp to each interaction ``in order to build time segments and clusters of nodes whose edge distributions are similar and evolve in the same way over the time segments''. In order to obtain the optimal number of time and nodes clusters, we maximize the integrated complete-data likelihood (ICL) using a greedy search in a very similar fashion as in Wyse, Frial and Latouche \cite{Latouche2}. This paper is structured as follows: in Section 2 we present the classical LBM and detail the time extension proposed. In Section 3 we derive the ICL for this model and in Section 4 we discuss the experiments we conducted with both simulated and real data. The Section 5 concludes the paper. 

\section{A non stationary latent block model}
We present here the LBM (Holland et al. \cite{Holland}), as described in Wyse, Friel, Latouche (2014). Two sets of nodes are considered: $A=\{a_1, \dots, a_{N} \}$ and $B=\{ b_1,\dots, b_{M}\}$.  Undirected links between node $i$ from $A$ and node $j$ from $B$, are counted by the observed variable $X_{ij}$, being the component $(i,j)$ of the $N \times M$ adjacency matrix $X=\{X_{ij}\}_{i \leq N, j \leq M}$. Nodes in $A$ and $B$ are clustered in $K$ and $G$ disjointed subgroups respectively: 
\begin{equation*}
 A= \cup_{k \leq K}A_k, \qquad A_i \cap A_j=\emptyset, \quad\forall i\neq j  
\end{equation*}
and similarly for $B$. Nodes in the same cluster in $A$ have linking attributes of the same nature to clusters of $B$. 
We introduce two hidden vectors $\mathbf{c}=\{c_1,\dots, c_N\}$ and $\mathbf{w}=\{w_1,\dots, w_M\}$ labeling each node's membership:
\begin{equation*}
 c_i=k\qquad\textit{iff}\quad{a_i \in A_k,\quad \forall k\leq K}\quad\text{and}\quad w_j=g\quad\textit{iff}\quad{b_j \in B_g,\quad \forall g\leq G}.
\end{equation*}
In order to introduce the temporal dimension, consider now a sequence of equally spaced, adjacent time steps $\{\Delta_u:= t_u - t_{u-1} \}_{u \leq U}$ over the interval $[0,T]$ and a partition $C_1, \dots, C_D$ of the same interval\footnote{$T$ and $U$ are linked by the following relation: $T=\Delta_u U$.}. We introduce furthermore a random vector $\mathbf{y}=\{y_u\}_{u \leq U}$, such that $y_u=d$ if and only if $I_u:=]t_{u-1}, t_{u}] \in C_d, \forall d \leq D$. 
We attach to $\mathbf{y}$ a multinomial distribution:
\begin{equation*}
 p(\mathbf{y} | \boldsymbol{\beta},D)=\prod_{d \leq D}\beta_d^{|C_d|},
\end{equation*}
where $|C_d|=\#\{I_u : I_u \in C_d\}$. Now we define $N^{I_u}_{ij}$ as the number of observed connections between $a_i$ and $b_j$, in the time interval $I_u$ and we make the following crucial assumption:
\begin{equation}
 p(N^{I_u}_{ij}| c_i=k, w_j=g, y_u=d)  \qquad\text{follows a}\qquad\text{Poisson}(\Delta_u\lambda_{kgd}),
\end{equation}
hence the number of interactions is \emph{conditionally} distributed like a Poisson random variable with parameter depending on $k,g,d$ ($\Delta_u$ is constant).


\medskip
\textbf{Notation}: In the following, for seek of simplicity, we will note:
\begin{equation*}
\prod_{k,g,d} := \prod_{k \leq K} \prod_{g \leq G}\prod_{ d \leq D} \quad\text{and}\quad \prod_{c_i}:= \prod_{i:c_i=k}
\end{equation*}
and similarly for $\prod_{w_j}$ and $\prod_{y_u}$.

\medskip
The adjacency matrix, noted $N^{\Delta}$, has three dimensions ($N\times M \times U$) and its observed likelihood can be computed explicitly:
\begin{equation}
\label{eq:obsLL}
 p(N^{\Delta}|\Lambda, \mathbf{c}, \mathbf{w}, \mathbf{y}, K,G,D)= \prod_{k,g,d} \frac{\Delta^{S_{kgd}}}{\prod_{c_i} \prod_{w_j}\prod_{y_u} N_{ij}^{I_u}! }e^{-\Delta \lambda_{kgd} R_{kgd}}\lambda_{kgd}^{S_{kgd}}, 
 \end{equation}
where we noted $S_{kgd}:= \sum_{c_i}\sum_{w_j} \sum_{y_u} N^{I_u}_{ij}$ and $R_{kgd}:=|A_k||B_g||C_d|$ and the subscript $u$ was removed from $\Delta_u$ to emphasize that time steps are equally spaced for every $u$.

Since $\mathbf{c}$,$\mathbf{w}$ and $\mathbf{y}$ are not known, a multinomial factorizing probability density $p(\mathbf{c}, \mathbf{w}, \mathbf{y}| \Phi, K,G,D)$, depending on hyperparameter $\Phi$, is introduced. The joint distribution of labels looks finally as follows:
 \begin{equation}
 \label{eq:term_1_ns}
   p(\mathbf{c}, \mathbf{w}, \mathbf{y} |\Phi, K, G, D) = \left(\prod_{k \leq K}\omega_k^{|A_k|}\right)\left(\prod_{g\leq G}\rho_g^{|B_g|}\right)\left(\prod_{d \leq D}\beta_d^{|C_d|}\right),
\end{equation}
where $\Phi=\{\boldsymbol{\omega}, \boldsymbol{\rho},\boldsymbol{\beta} \}$.
\section{Exact ICL for non stationary LBM}
\subsection{Exact ICL derivation}
The integrated classification criterion (ICL) was introduced as a model selection criterion in the context of Gaussian mixture models by Biernacky et al. \cite{Biernacky}. C\^{o}me and Latouche \cite{Latouche1} proposed an exact version of the ICL based on a Bayesian approach for the stochastic block model and Wyse, Friel and Latouche \cite{Latouche2} applied the exact ICL to select the number of clusters in a bipartite network using an LBM model. This is the approach we follow here. 
The quantity we focus on is the \emph{complete data} log-likelihood, integrated with respect to the model parameters $\Phi$ and $\Lambda=\{\lambda_{kgd}\}_{k\leq K, g \leq G, d\leq D}$:
\begin{equation}
 \mathcal{ICL}= \log\left( \int p(N^{\Delta},\mathbf{c}, \mathbf{w}, \mathbf{y}, \Lambda, \Phi| K,G,D)d\Lambda d\Phi \right).
\end{equation}
Introducing a prior distribution $\nu(\Lambda, \Phi | K,G,D)$ over the pair $\Phi, \Lambda$ and thanks to ad hoc independence assumptions,
the ICL can be rewritten as follows:
\begin{equation}
\label{eq: ICLNI}
 \mathcal{ICL}= \log\left(\nu(N^{\Delta}| \mathbf{c}, \mathbf{w}, \mathbf{y},K,G,D)\right) + \log\left( \nu(\mathbf{c}, \mathbf{w}, \mathbf{y}| K,G,D)\right).
\end{equation}
The choice of prior distributions over the model parameters is crucial to have an explicit form of the ICL.

\subsection{A priori distributions}
We consider the conjugate prior distributions. Thus we impose a Gamma a priori over $\Lambda$:
\begin{equation*}
  \nu(\lambda_{kgd}| a_{kgd}, b_{kgd})=\frac{b_{kgd}^{a_{kgd}}}{\Gamma(a_{kgd})}\lambda_{kgd}^{a_{kgd}-1}e^{-b_{kgd}\lambda_{kgd}}
\end{equation*}
and a factorizing Dirichlet a priori distribution to $\Phi$:
\begin{equation*}
 \nu(\Phi|K,G,D) = \text{Dir}_K(\boldsymbol{\omega}; \alpha,\dots,\alpha) \times \text{Dir}_G(\boldsymbol{\rho}; \delta,\dots,\delta) \times \text{Dir}_D(\boldsymbol{\beta}; \gamma,\dots,\gamma).
\end{equation*}
It can be proven  that the two terms in \eqref{eq: ICLNI}, reduce to:
\begin{align}
 \label{eq:ICLI1}
   \nu(N^{\Delta}| \mathbf{c}, \mathbf{w}, \mathbf{y},K,G,D) =&  \prod_{k,g,d} \frac{b_{kgd}^{a_{kgd}}\Delta^{S_{kgd}}}{\Gamma{(a_{kgd}}) \prod_{c_i} \prod_{w_j}\prod_{y_u} N_{ij}^{I_u}! }\\
   &\frac{\Gamma(S_{kgd}+a_{kgd})}{[\Delta R_{kgd}+ b_{kgd}]^{S_{kgd} + a_{kgd}}} \nonumber
\end{align}
and:
\begin{align}
 \nu(\mathbf{c}, \mathbf{w}, \mathbf{y}| K,G,D)=&\frac{\Gamma(\alpha K)}{\Gamma(\alpha)^K}\frac{\prod_{k\leq K}\Gamma(|A_k| + \alpha)}{ \Gamma(N + \alpha K)}
\times \frac{\Gamma(\delta G)}{\Gamma(\delta)^G}\frac{\prod_{g\leq G}\Gamma(|B_g| + \delta)}{ \Gamma(M + \delta G)} \nonumber \\
 \times& \frac{\Gamma(\gamma D)}{\Gamma(\gamma)^D}\frac{\prod_{d\leq D}\Gamma(|C_d| + \gamma)}{ \Gamma(U + \gamma D)}.
 \end{align}

\subsection{ICL Maximization}

In order to maximize the integrated complete likelihood (ICL) in equation \eqref{eq: ICLNI} with respect to the six unknowns $\mathbf{c}, \mathbf{w}, \mathbf{y}, K, G, D$, we rely on a greedy search over labels and the number of nodes and time clusters. This approach is described in Wyse, Frial and Latouche \cite{Latouche2} for a stationary latent block model.     

\section{Experiments}
\subsection{Simulated data}

Some experiments on simulated data have initially been conducted. Based on the model described in Section 2, we simulated interactions between 50 \emph{source} nodes and 50 \emph{destination} nodes, both clustered in three groups ($K,G=3$). Interactions take place into 24 time intervals of unitary length (ideally one hour), clustered into three groups too ($D=3$). 
Nodes and time intervals labels are sampled from multinomial distributions, whose hyperparameters $(\boldsymbol{\omega}, \boldsymbol{\rho}, \boldsymbol{\delta})$ have all been set equal to $\{1/3, 1/3, 1/3  \}$. 
With these settings, we consider 27 different Poisson parameters ($\lambda$s) generating connections between nodes. The generative model used to produce them is described by:

\begin{equation*}
  \lambda_{kgl}= s_1[k] + s_2[g] + s_3[l], \qquad\ k,g,l \in \{1, 2, 3\}
\end{equation*}
where:

\begin{equation*}
 s_1=[0,2,4] \quad s_2=[0.5,1,1.5] \quad s_3=[0.5,1,1.5]
\end{equation*}
ans $s_1[k]$ denotes the k-th component of $s_1$. Similarly for $s_2$ and $s_3$. 
The greedy search algorithm we coded was able to exactly recover these initial settings, converging to the true ICL of $-122410$. Other experiments were run with different values inside vectors $s_1, s_2, s_3$. Not surprisingly  the more nuanced differences between $\lambda$s are, the more difficult it is for the algorithm to converge to the true value of the ICL\footnote{Greedy search algorithms are path dependent and they could converge to local maxima.}.

\subsection{Real Data}

The dataset we used was collected during the \textbf{ACM Hypertext} conference held in Turin, June 29th - July 1st, 2009. Conference attendees volunteered to to wear radio badges that monitored their face-to-face proximity. The dataset represents the dynamical network of face-to-face proximity of 113 conference attendees over about 2.5 days\footnote{More informations can be found at: 

\url{http://www.sociopatterns.org/datasets/hypertext-2009-dynamic-contact-network/ }.

}. Further details can be found in Isella, Stehl\'e, Barrat, Cattuto, Pinton, Van den Broeck \cite{Cattuto}. We focused on the first conference day, namely the twenty four hours going from 8am of June 29th to 7.59am of June 30th. The day was partitioned in small time intervals of 20 seconds in the original data frame.  We considered 15 minutes time aggregations, thus leading to a partition of the day made of 96 consecutive quarter-hours ($U=96$ with previous notation). A typical row of the adjacency matrix we analyzed, looks like:

\begin{center}
 \begin{tabular}{c|c|c|c}
  \hline
  \footnotesize\emph{Person 1} & \footnotesize\emph{Person 2} & \footnotesize\emph{Time Interval (15m)} & \footnotesize\emph{Number of interactions} \\
  \hline
   52 & 26 & 5 & 16 \\
  \hline 
 \end{tabular}
\end{center}
It means that conference attendees 52 and 26, between 9am and 9.15am have spoken for $16 \times 20s \approx 5m30s$.  

The greedy search algorithm converged to a final ICL of -53217.4, corresponding to 23 clusters for nodes (people) and 3 time clusters. 
\begin{figure}[!h]
\centering
\begin{subfigure}{.5\textwidth}
  \centering
  \includegraphics[width=\linewidth]{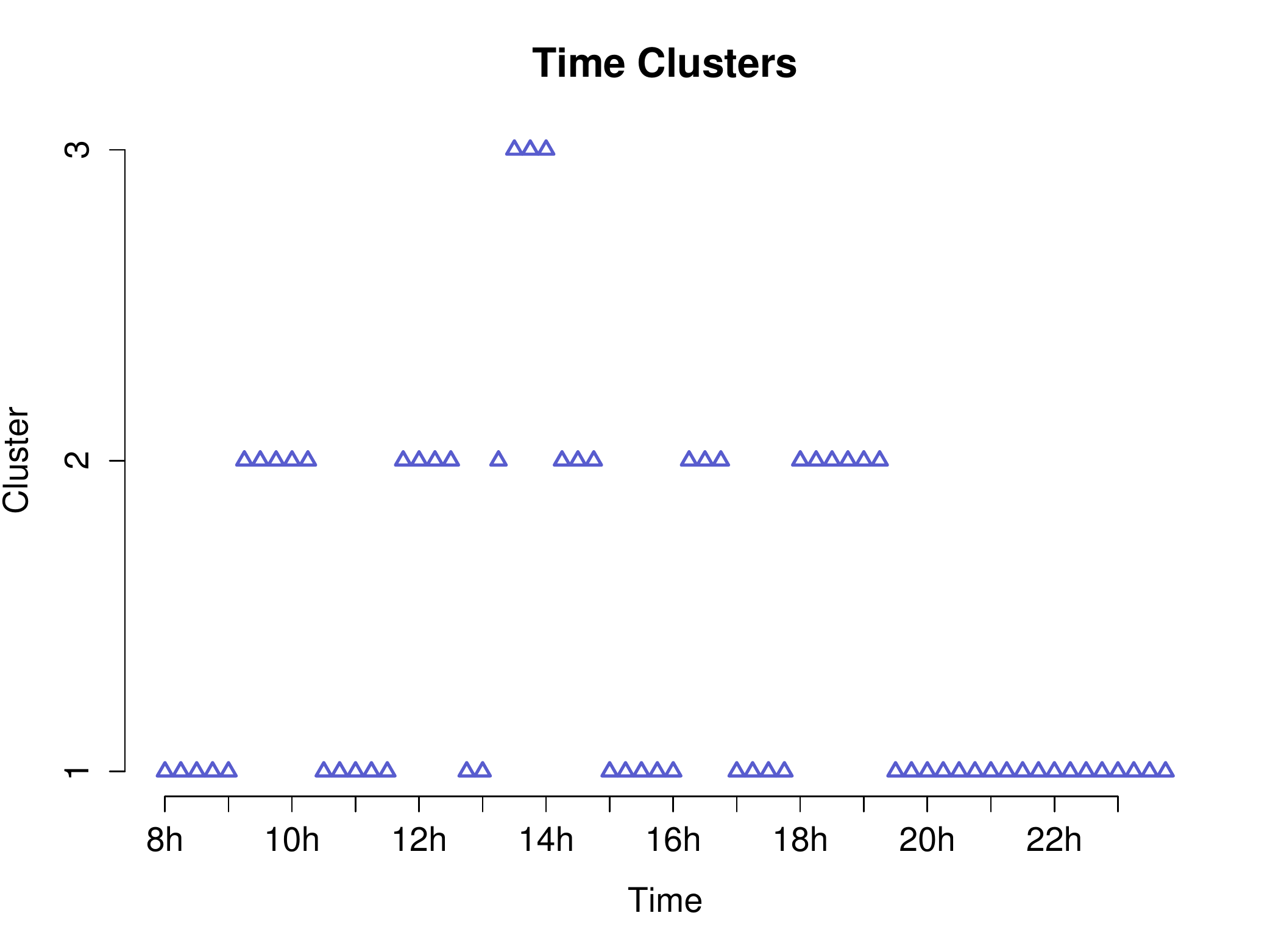}
  \caption{\footnotesize Clustered time intervals.}
  \label{fig:TC}
\end{subfigure}%
\begin{subfigure}{.5\textwidth}
  \centering
  \includegraphics[width=\linewidth]{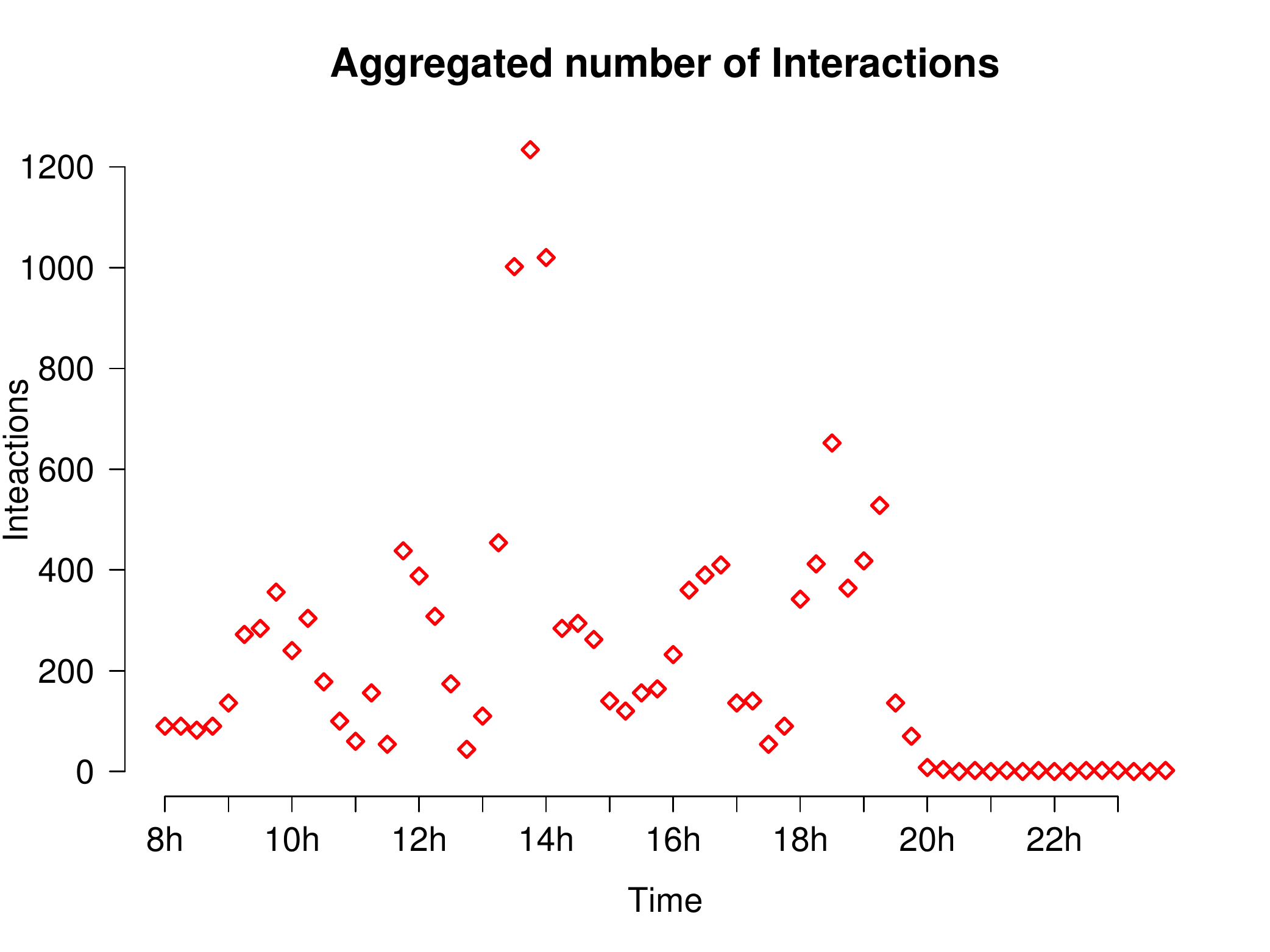}
  \caption{\footnotesize Connections for every time interval.}
  \label{fig:Agginter}
\end{subfigure}
\caption{\small The aggregated connections for every time interval (\ref{fig:Agginter}) and time clusters found by our model (\ref{fig:TC}) are compared.}
\label{fig:MainFIgure}
\end{figure}
In Figure \eqref{fig:TC} we show how daily quarter-hours are assigned to each cluster: the class $C_1$ contains intervals marked by a weaker intensity of interactions (on average), whereas  intervals inside $C_3$ are characterized by the highest intensity of interactions. This can either be verified analytically by averaging estimated Poisson intensities for each one of the three clusters or graphically by looking at Figure \eqref{fig:Agginter}. In this Figure we computed the total number of interactions between conference attendees for each quarter-hour and it can clearly be seen how time intervals corresponding to the higher number of interactions have been placed in cluster $C_3$, those corresponding to an intermediate interaction intensity, in $C_2$ and so on. It is interesting to remark how the model can quite closely recover times of social gathering like the lunch break (13.00-15.00) or the ``wine and cheese reception'' (18.00-19.00). 
A complete program of the day can be found at:

\url{http://www.ht2009.org/program.php}.

\section{Conclusions}

We proposed a non-stationary evolution of the latent block model (LBM) allowing us to simultaneously infer the time structure of a bipartite network  and cluster the two node sets. The approach we chose consists in partitioning the entire time horizon in fixed-length time intervals to be clustered on the basis of the intensity of connections in each interval. We derived the complete ICL for such a model and maximized it numerically, by means of a greedy search, for two different networks: a simulated and a real one. The results of these two tests highlight the capacity of the model to capture non-stationary time structures. 


\begin{footnotesize}





\end{footnotesize}


\end{document}